\documentclass[conference]{IEEEtran}
\IEEEoverridecommandlockouts
\usepackage{cite}
\usepackage{amsmath,amssymb,amsfonts}
\usepackage{algorithmic}
\usepackage{graphicx}
\usepackage{textcomp}
\usepackage{xcolor}
\usepackage{babel}
\def\BibTeX{{\rm B\kern-.05em{\sc i\kern-.025em b}\kern-.08em
    T\kern-.1667em\lower.7ex\hbox{E}\kern-.125emX}}
\begin{document}

\title{Character-Level Bangla Text-to-IPA Transcription Using Transformer Architecture with Sequence Alignment}

\author{\IEEEauthorblockN{Jakir Hasan}
\IEEEauthorblockA{\textit{Computer Science and Engineering} \\
Shahjalal University of Science\\
and Technology\\
Sylhet, Bangladesh \\
jakirhasan718@gmail.com}

\and
\IEEEauthorblockN{Shrestha Datta}
\IEEEauthorblockA{\textit{Computer Science and Engineering} \\
Shahjalal University of Science\\
and Technology\\
Sylhet, Bangladesh \\
shresthadatta910@gmail}
\and
\IEEEauthorblockN{Ameya Debnath}
\IEEEauthorblockA{\textit{Computer Science and Engineering} \\
Shahjalal University of Science\\
and Technology\\
Sylhet, Bangladesh \\
ameyadebnath@gmail.com}
}

\maketitle

\begin{abstract}
The International Phonetic Alphabet (IPA) is indispensable in language learning and understanding, aiding users in accurate pronunciation and comprehension. Additionally, it plays a pivotal role in speech therapy, linguistic research, accurate transliteration, and the development of text-to-speech systems, making it an essential tool across diverse fields. Bangla being 7th as one of the widely used languages, gives rise to the need for IPA in its domain. Its IPA mapping is too diverse to be captured manually giving the need for Artificial Intelligence and Machine Learning in this field. In this study, we have utilized a transformer-based sequence-to-sequence model at the letter and symbol level to get the IPA of each Bangla word as the variation of IPA in association of different words is almost null. Our transformer model only consisted of 8.5 million parameters with only a single decoder and encoder layer. Additionally, to handle the punctuation marks and the occurrence of foreign languages in the text, we have utilized manual mapping as the model won't be able to learn to separate them from Bangla words while decreasing our required computational resources. Finally, maintaining the relative position of the sentence component IPAs and generation of the combined IPA has led us to achieve the top position with a word error rate of 0.10582 in the public ranking of DataVerse Challenge - ITVerse 2023 (https://www.kaggle.com/competitions/dataverse\_2023/).
\end{abstract}


\section{Introduction}
The International Phonetic Alphabet (IPA) stands as a standardized notation system of great significance in the realms of linguistic inquiry and language acquisition. Through its extensive array of symbols representing speech sounds, the IPA greatly facilitates precise pronunciation and comprehension, benefiting learners, linguists, and therapists alike. Its impact also extends across a range of disciplines, encompassing linguistics research, speech therapy, precise transliteration, and the advancement of cutting-edge speech synthesis systems.

Within the domain of Bengali, a language counted among the top seven most widely spoken worldwide, the imperative for a robust IPA system becomes apparent. The intricate phonetic intricacies of Bengali demand a tailored approach to IPA generation. Present endeavors in this direction remain constrained, underscoring the need for the integration of advanced AI and ML methodologies to bridge this gap.

IPA serves various purposes. Initially, it is employed in contemporary French and English dictionaries, as well as pronunciation guides, to accurately represent the pronunciation of words \cite{pierret1994phonetique}. Additionally, the IPA can provide a foundation for creating a written system for languages that have yet to be standardized in writing \cite{lambert1997}. Nowadays, linguists and speech-language pathologists (SLPs) make more frequent use of the IPA (Munot and Nève 2002). Linguists utilize it to gain intricate insights into the mechanics of languages and their evolutionary processes. Although there has been no previous work related to IPA in the Bangla domain first work was related to Bangla pronunciation utilizing an encoder-decoder with Gated Recurrent Unit – Recurrent Neural Networks (GRU-RNNs) \cite{ahmad2018sequence}.

The DataVerse Challenge - ITVerse 2023 \cite{dataverse2023} presented the unique dataset of Bangla text to IPA with the chance to explore and solve the challenges of IPA generation from Bangla text. This challenge also provides first-hand experience to explore and experience the dataset and possible solutions with Machine Learning (ML) and Artificial Intelligence (AI) to generate IPA of corresponding Bangla texts. This takes us one step closer to creating more refined open-source text-to-speech synthesizer for one of the most popular language, Bangla.

As the IPA generation of Bangla text can be viewed as a similar task to language translation where we are translating Bangla to an IPA representation. With the introduction of attention-based transformers \cite{vaswani2017attention}, they have shown noticeable performance in a wide variety of tasks. Transformers have shown a remarkable performance in the domain of machine translation tasks \cite{wang2019learning}. Hence we have approached the solution to this problem with transformer architecture.

\section{Dataset Analysis}
\begin{figure}
    \centering
    \includegraphics[width=0.5\textwidth]{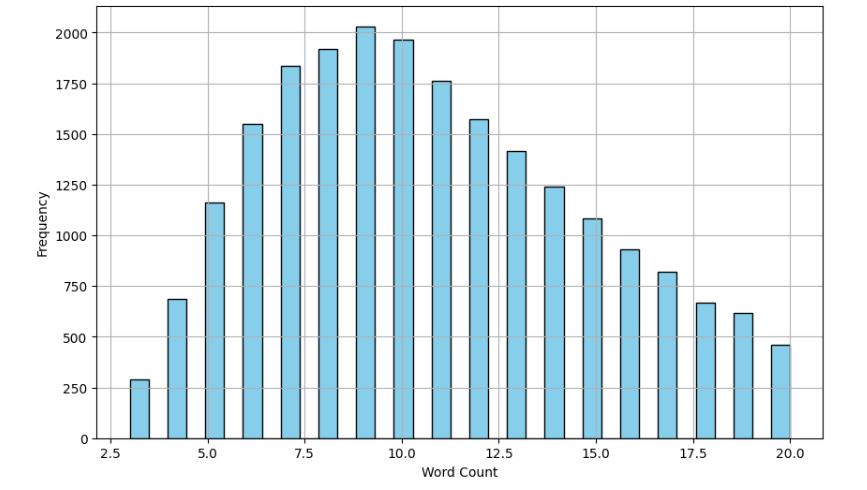}
    \caption{Word count histogram of training dataset}
    \label{fig:fig_1}
\end{figure}

In the DataVerse Challenge - ITVerse 2023, two training datasets were provided. We worked with the first training dataset. Our training dataset consisted of 21999 samples where Bengali text and their corresponding IPA transcription were provided. The test dataset consisted of 27228 samples with Bengali texts. The number of test samples is approximately 1.24 times that of the number of training samples. The number of unique tokens in training and test samples are 37807 and 48053 respectively. Figure \ref{fig:fig_1} shows the word count histogram of our training dataset.

In the text column of the training dataset, the number of unique characters is 136 and for the IPA column, it is 56.
After carefully observing the character set of both the text and ipa column we noticed that there were no samples in the ipa column where Bengali words are present in IPA transcription but there were 19 samples in the text column where English words were present in Bengali sentences. Furthermore, there were 7 samples where the number of words in the text and their corresponding IPA were different.

In the training dataset, the maximum length of a word in the text column is 31, and in the ipa column, it is 34. The maximum length of a word in the text column of the test dataset is 36. The training dataset doesn't contain any words with Bengali numerals. But in the test data, 9602 samples have Bengali numerals which is almost 30\% of the test samples.

\section{Methodologies}
This section discusses the methodologies we have adopted in applying our system for the contest.
\subsection{Training Data Preparation and Preprocessing}
From the dataset, it is noticeable that the IPA of each word is not mostly affected by their neighboring words. Leading to the scope of translating one word at a time which reduces the model's computational resources. Hence we took the Bangla words from the dataset and their corresponding IPA leading to our training set consisting of each sample being Bangla word and corresponding IPA. We excluded any repetitive occurrence of the same word. Our goal was to generate IPA transcription of a word at the character level. So, we inserted spaces on neighboring characters of both Bengali words and their corresponding IPA for generating sequences of characters without changing their relative positions. We have built the training and validation dataset from these sequence pairs making the size of our training dataset 37807.

\subsection{Handling Punctuation Marks and Special Symbols}
The punctuation marks and related special symbols were not changed and held their relative positions in the IPA of Bangla sentences. As handling these manually is more efficient than training the transformer model to learn to differentiate letters from various punctuation marks, we kept the punctuation marks unchanged along with their relative position in the resulting IPA.

\subsection{Handling Foreign language Words and Characters}
As the main focus of the dataset is Bangla IPA, it does not present the IPA of foreign language. Hence before feeding our trained model, we excluded foreign language words and characters, thus returning an empty IPA for each occurrence of foreign language words and characters.

\subsection{Transformer Model}
The IPA construction mainly depends on the position and association of various letters. Also to generate the IPA we need to generate letters instead of whole words to cope with unseen data. This gives the necessity for the vocabulary to consist of letters instead of words or a collection of letters. Hence the text vectorizer consisted of letter vocabulary with vocabulary adaptation enabled. We have used encoder-decoder based transformer model for this task with a single encoder and a single decoder. This model consists of about 8.5 million parameters. The model was trained with the root mean square propagation optimization algorithm utilizing sparse categorical cross-entropy loss function while limiting the max input output length to 64.

\subsection{IPA Dictionary and Bangla-IPA mapping}
We have mapped the IPA for corresponding Bangla words. As same words may be repeated multiple times leading to the prediction of the same IPA by the model decreasing computational resource efficiency. Hence we kept an IPA dictionary for mapping the Bangla words to their respective IPA. This dictionary records a word and its corresponding IPA whenever a new word is faced and its IPA is generated. As the search time of words in a dictionary is logarithmic word length time complexity, it is more time efficient to use a dictionary instead of generating IPA for previously encountered words again and again.

\section{Experimental Setup}
In this section, we will talk more vigorously regarding setting up our experiment, the process of training, and the process of inference on the contest test set.
\subsection{Model Training}
We have split our total training data into 90:5:5 ratios as training, validation, and test sets for the holdout method of model validation. We took accuracy as our validation metric. This led to achieving the highest validation score of 98\% accuracy. Instead of selecting hyperparameters of the best performing model, we took hyperparameters which had stable performance close to the best performing model. After selecting proper hyperparameters we trained our final model on 99\% of training data and validated on the rest. For inference optimization, we created an IPA dictionary with unique words and their corresponding IPA during model training. We trained the model for 50 epochs with a learning rate of 0.001 and batch size of 64 with mini-batch approach.

\subsection{Inference}
\begin{figure}
    \centering
    \includegraphics[width=0.5\textwidth]{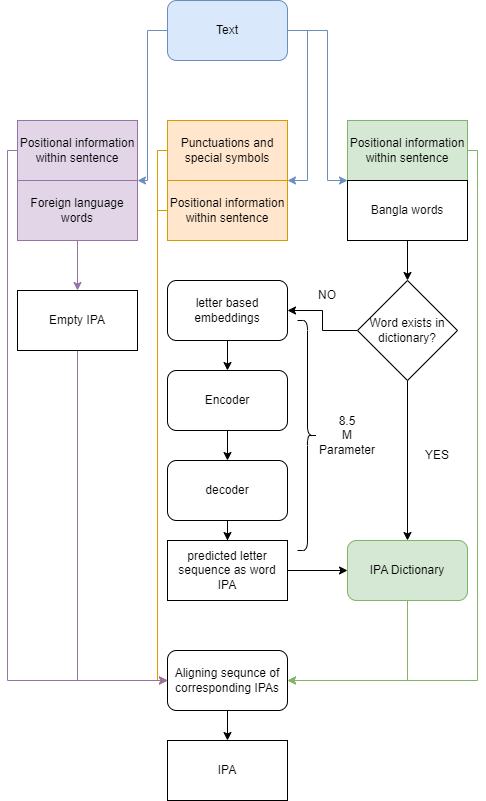}
    \caption{Architechture of our system}
    \label{fig:fig_2}
\end{figure}

For each Bangla text sample from our contest test data, we kept the punctuations unchanged, generated empty IPA corresponding to foreign words, and generated IPA with our trained transformer model for the Bangla words. We built the whole IPA of the sample text, substituting each text component with the resultant IPA keeping the relative position unchanged, i.e. aligning the sequence. To further optimize the process of IPA generation, we updated the IPA dictionary with each new unique word and its corresponding generated IPA. Whenever the same word is found for the second time in the inference process we fetch the IPA from the dictionary instead of using the trained model for generating the same IPA. Our system architecture \ref{fig:fig_2} presents the visual relationship among different components of the system.

\section{Result and Analysis}
It is to be noted that, all the results mentioned here are based on the public leaderboard which is based on 52\% of test data. All of our experimental results are provided in this section.
We have carried out our experiments in Kaggle \cite{kaggle} notebooks using
GPU T4 x2 as runtime. We have used Tensorflow 2.12.0 \cite{tensorflow} and Python
3.10.12 in our experiments.

After training the model is saved. The size of our test dataset is 27228. Word Error Rate (WER) is used for evaluating the performance of our model. We have taken iterative approaches to solve different Bangla language-specific tasks to gain lower word error rates and thus improve the robustness of our model. Table \ref{accuracyresult} shows the word error rates after solving different subtasks for Bangla text to IPA transcription. 

\subsection{Direct Mapping}
In our first approach, after careful analysis of the training dataset, we observed that in most of the cases most of the cases the number of words in Bangla text is the same as its corresponding IPA transcription. After handling mismatched sequence lengths, we mapped every Bangla word with its IPA. Our first model was trained on these sequence pairs on character level and word error rate was 20.48\%. This indicates that our trained model for IPA mapping predicts most of the Bangla non-numerical texts correctly leaving only to explore the rest about 20\% of errors.

\subsection{Punctuations and English Letters Handling}
\label{punceng}
In the IPA transcription of Bengali text, English words are ignored and punctuations are kept in place. So, we handled these in the preprocessing steps and our achieved word error rate was 11.41\%. This significant boost to performance highlights the importance of correct relative positions of the IPA and the importance of handling non-conventional textual properties for IPA generation.

\subsection{Bengali Numerals Handling}
A large number of test data had Bengali numerals. Our previous two approaches didn't handle them and simply kept them as it is. In this approach, we converted every numeral into its word representation and predicted its IPA. But in this case, our word error rate increased to 13.43\%. This is because a single numeral can generate many words and that's why the effect of any incorrect single word to IPA transcription can decrease the model's performance significantly.

For further intuition into this matter, we tried representing only single-digit numerals to words. Surprisingly, this didn't change the performance leading to the intuition that, either there was no IPA for Bangla numerals or the Bangla numerals are pronounced differently compared to their written form. As the training set did not contain any IPA related to numerals it is difficult to come to a concrete conclusion. This will be further analyzed in \ref{remfor}.

\subsection{All Foreign Symbols Handling}
\label{remfor}
In our previous section, we mentioned that substituting Bengali numerals with words and creating their transcription decreases our model's accuracy. That's why in this approach we substitute every numeral with an empty string. We observed that the training set has a lot of unknown symbols that are not used in Bengali and our test might contain foreign characters also. In order to handle foreign characters we build up a vocabulary of all possible Bengali characters and symbols and eliminate the foreign characters and unnecessary symbols. Also in the post-processing step, we handle the case when our model's predicted output is null. After handling these issues we achieved a word error rate of 10.58\%.

This improvement backs up the improvement of section \ref{punceng}. Interestingly, removing the Bangla numerals from the IPA did not change the performance leading to the conclusion that each Bangla numeral has a valid IPA and the IPA for word representation of Bangla numerals does not follow the same rule as the IPA representation of normal Bangla Words. This takes us to the view that it would be better to have a more in depth study with Bangla numeral IPA.


\begin{table}[htbp] 
\caption{Word Error Rates After Solving Subtasks}
\centering
\resizebox{\columnwidth}{!}{%
\begin{tabular}{|c|c|}
\hline
Approach & Word Error Rate (\%)\\
\hline
Direct mapping & 20.48 \\
\hline
Punctuations and English letters handling & 11.41  \\
\hline
Bengali numerals handling & 13.43  \\
\hline
All foreign symbols handling & 10.58  \\
\hline
\end{tabular}%
}
\label{accuracyresult}
\end{table}

\section{Conclusion}
In conclusion, we can say that Bangla IPA is mostly independent of its associative IPAs letting us achieve a satisfactory performance by generating wordwise IPA. The challenge of Bangla text to IPA can be easily handled by the sequence-to-sequence encoder-decoder based transformer architecture with letter-based vocabulary. Another of our key findings is that Bangla numerals IPA doesn't follow the same rule of conversion from its related word representation i.e. different pronunciation rules compared to general Bangla word pronunciations. This leads to the need for inclusion as much as variations of Bangla numerical IPAs in the dataset. Moreover, main optimization of such words to IPA systems is the inclusion of a dictionary, which keeps track of word-IPA pairs, as the same words result in the same IPA most of the time. Handling word-by-word IPA also decreases the model hassle for handling special symbols and foreign languages. As they can be handled easily externally without the need for complex ML models, boosted performance and efficient resource utilization are added to the IPA generation system. These methods enabled us to reduce our transformer model to only 8.5 million parameters with only one encoder and decoder layer.

\vspace{12pt}

\end{document}